\documentclass{article}

\usepackage{microtype}
\usepackage{graphicx}
\usepackage{booktabs} % for professional tables
\usepackage{multirow}
\usepackage{hyperref}
\usepackage{xcolor}
\usepackage{stmaryrd}

\usepackage[accepted]{icml2025}

\usepackage{amsmath}
\usepackage{amssymb}
\usepackage{mathtools}
\usepackage{amsthm}
\usepackage[scr=rsfs]{mathalpha}
\usepackage{subcaption}

\usepackage[capitalize,noabbrev]{cleveref}

\theoremstyle{plain}

\theoremstyle{definition}

\theoremstyle{remark}

\usepackage[textsize=tiny]{todonotes}

\icmltitlerunning{Maximize Your Diffusion: A Study into Reward Maximization and Alignment for Diffusion-based Control}

\begin{document}

\twocolumn[
\icmltitle{Maximize Your Diffusion: A Study into Reward Maximization \\and Alignment for Diffusion-based Control.}

% It is OKAY to include author information, even for blind
% submissions: the style file will automatically remove it for you
% unless you've provided the [accepted] option to the icml2025
% package.

% List of affiliations: The first argument should be a (short)
% identifier you will use later to specify author affiliations
% Academic affiliations should list Department, University, City, Region, Country
% Industry affiliations should list Company, City, Region, Country

% You can specify symbols, otherwise they are numbered in order.
% Ideally, you should not use this facility. Affiliations will be numbered
% in order of appearance and this is the preferred way.
% \icmlsetsymbol{equal}{*}

\begin{icmlauthorlist}
\icmlauthor{Dom Huh}{yyy}
\icmlauthor{Prasant Mohapatra}{yyy,comp}
% \icmlauthor{Firstname3 Lastname3}{comp}
% \icmlauthor{Firstname4 Lastname4}{sch}
% \icmlauthor{Firstname5 Lastname5}{yyy}
% \icmlauthor{Firstname6 Lastname6}{sch,yyy,comp}
% \icmlauthor{Firstname7 Lastname7}{comp}
% %\icmlauthor{}{sch}
% \icmlauthor{Firstname8 Lastname8}{sch}
% \icmlauthor{Firstname8 Lastname8}{yyy,comp}
% %\icmlauthor{}{sch}
% %\icmlauthor{}{sch}
\end{icmlauthorlist}

\icmlaffiliation{yyy}{Department of Computer Science, University of California, Davis, USA}
\icmlaffiliation{comp}{University of South Florida, USA}
% \icmlaffiliation{sch}{School of ZZZ, Institute of WWW, Location, Country}

\icmlcorrespondingauthor{Dom Huh}{dhuh@ucdavis.edu}
% \icmlcorrespondingauthor{Firstname2 Lastname2}{first2.last2@www.uk}

% You may provide any keywords that you
% find helpful for describing your paper; these are used to populate
% the "keywords" metadata in the PDF but will not be shown in the document
\icmlkeywords{®ate Optimization}

\vskip 0.3in
]

% this must go after the closing bracket ] following \twocolumn[ ...

% This command actually creates the footnote in the first column
% listing the affiliations and the copyright notice.
% The command takes one argument, which is text to display at the start of the footnote.
% The \icmlEqualContribution command is standard text for equal contribution.
% Remove it (just {}) if you do not need this facility.

\printAffiliationsAndNotice{}  % leave blank if no need to mention equal contribution
% \printAffiliationsAndNotice{\icmlEqualContribution} % otherwise use the standard text.

\begin{abstract}
Diffusion-based planning, learning, and control methods present a promising branch of powerful and expressive decision-making solutions. Given the growing interest, such methods have undergone numerous refinements over the past years. 
However, despite these advancements, existing methods are limited in their investigations regarding general methods for reward maximization within the decision-making process. 
In this work, we study extensions of fine-tuning approaches for control applications. 
Specifically, we explore extensions and various design choices for four fine-tuning approaches: reward alignment through reinforcement learning, direct preference optimization, supervised fine-tuning, and cascading diffusion. We optimize their usage to merge these independent efforts into one unified paradigm. 
We show the utility of such propositions in offline RL settings and demonstrate empirical improvements over a rich array of control tasks.
\end{abstract}

\section{Introduction}
Denoising diffusion probabilistic models (DDPM) have demonstrated incredible successes in numerous domains as one of the de facto standards for generative modeling in continuous domains~\citep{sohl2015deep, ho2020denoisingdiffusionprobabilisticmodels, yang2024diffusionmodelscomprehensivesurvey}, garnering heightened attention in research communities with the model's exceptional ability to capture highly complex data distributions and flexibility in constraint-based optimization that can be applied to diverse modalities. A notable direction is their adoption and realizations in decision-making applications~\citep{zhu2024diffusionmodelsreinforcementlearning}, contributing to an evolving class of model-based planning as well as model-free algorithms we will refer to as diffusion-based planning, learning, and control. Despite the appealing promises of DDPMs, the algorithm's sample generation inefficiencies loom as a notorious and significant concern~\citep{cao2024survey}, causing a severe bottleneck as an efficient solution for closed-loop control.

\begin{figure}
    \centering
    \includegraphics[width=\linewidth]{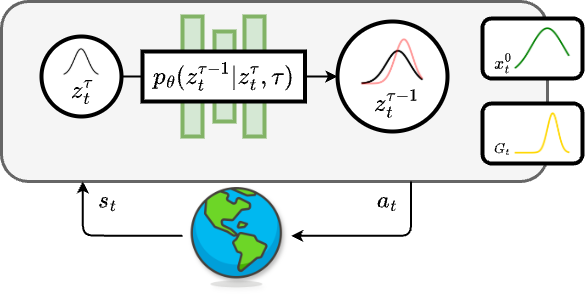}
    \caption{A illustration of the reward maximization of diffusion model for control, visualizing the diffusion process at $\tau \shortrightarrow \tau - 1$ such that $z^{\tau - 1}_t$ is denoised towards the target distribution $x^0_t$ but also aligned with maximize the return distribution $G_t$, shown in red.}
    \label{fig:enter-label}
\end{figure}

The focus of this work addresses this limitation, proposing a general investigation into reward maximization, or more aptly called \textit{reward alignment}, directly on the diffusion process for control. In turn, our objective is to reduce the \textit{number of samples} required to generate to maximize the reward signal. We adopt modern optimization procedures to facilitate a guided and constrained distributional shift biased towards maximal returns. Specifically, we investigate four reward alignment approaches --- reward maximization with reinforcement learning, direct preference optimization, supervised fine-tuning, and cascading diffusion --- and compare their performance in various control settings, more so in offline RL settings. We evaluate various design choices as well as an unification of the studied alignment approaches on a diverse testbed of challenging control tasks consisting of a variety of complex tasks from benchmarks such as D4RL \cite{fu2020d4rl}, MetaWorld \cite{yu2021metaworldbenchmarkevaluationmultitask}, PushT \cite{florence2022implicit}, Relay-Kitchen \cite{gupta2019relaypolicylearningsolving}, and RoboMimic \cite{mandlekar2021matters} as well as a novel evaluation task that enables in-depth analysis and visualization of the various features of the environment, the controller and their interactions.

\section{Background}

\paragraph{Diffusion Models}
DDPMs~\citep{ho2020denoisingdiffusionprobabilisticmodels} are commonly known as a class of generative models, which aim to estimate and sample from unknown data distributions $X$. DDPMs define a Markov chain of forward Gaussian kernels $q(x_t \vert x_{t-1})$ with a scheduled~\citep{nichol2021improved} sequence of variance coefficients $\{\beta_t \in (0,1)\}_{t=1}^T$ that gradually add noise to input data
\begin{equation}
    q(x_t \vert x_{t-1}) = \mathcal{N}(x_t; \sqrt{1-\beta_t} x_{t-1}, \beta_t \mathrm{I})
\end{equation}
and then learn to reverse the diffusion process by denoising with learnable reverse Gaussian kernels $p_\theta(z_{t-1} \vert z_t)$, parameterized by $\theta$, to generate data samples $z_0$ from a standard Gaussian noise prior $z_T \sim \mathcal{N}(0,I)$.
\begin{equation}
p_\theta(z_{t-1} \vert z_t) = \mathcal{N}(z_{t-1}; \mu_\theta(z_t, t), \Sigma(z_t, t))
\end{equation}
A common practice is to parameterize ${\mu}_\theta({z}_t, t)$ to perform $\epsilon-$prediction, motivated by its resemblance to the Langevin dynamics, which enables a simplification of the training objective.
\begin{equation}
    {\mu}_\theta({z}_t, t) \coloneqq {\frac{1}{\sqrt{\alpha_t}} \Big( {z}_t - \frac{1 - \alpha_t}{\sqrt{1 - \bar{\alpha}_t}} {\epsilon}_\theta({z}_t, t) \Big)}
\end{equation}
The training objective of DDPM uses the variational lower bounds of the marginal likelihood to minimize the negative log-likelihood~\citep{kingma2021variational}, as shown in Equation \ref{equation:nll}, where in practice, a simplified yet equivalent objective $\mathcal{L}_{\text{DDPM}}(\theta)$ is commonly used~\citep{ho2020denoisingdiffusionprobabilisticmodels}.
\begin{equation}\label{equation:nll}
    \mathcal{L}_{\text{NLL}}(\theta) = \mathbb{E}[-\log{p_\theta(z_0)}] \leq - \mathbb{E}_{q} [\log \frac{p_\theta({z}_{0:T})}{q({x}_{1:T} \vert {x}_{0})} ]
\end{equation}
\begin{align}
    \mathcal{L}_{\text{DDPM}}(\theta) = \mathbb{E}_{x_0}\Big[\|{\epsilon}_t - {\epsilon}_\theta(\sqrt{\bar{\alpha}_t}{x}_0 + \sqrt{1 - \bar{\alpha}_t}{\epsilon}_t, t)\|^2 \Big]
\end{align}
where $t\sim[0,T]$ with $T$ is the total number of diffusion steps, $\beta_t = 1 - \alpha_t$, $\bar{\alpha}_t = \mathop{\Pi}_{i=1}^t \alpha_i$, and $\epsilon_t \sim \mathcal{N}(0,\mathrm{I})$. A common and useful intuition is that DDPMs learn information regarding a time-dependent vector field $u_t(z_t, \theta)$ that generates a probability density/diffusion path $p_t(z_{t-1}|z_{t})$ from a simple and known distribution (e.g., standard Gaussian) to the unknown target distribution (e.g., data distribution).

Several recent improvements and reformulations of this DDPM paradigm have been explored~\citep{cao2024survey}, most notably and relevantly, towards efficient sample generation with implicit probabilistic models (DDIM~\citep{song2020denoising}), $v-$prediction~\citep{salimans2022progressive} and rectification~\citep{liu2022flow, wang2024rectified}, conditional guidance~\citep{dhariwal2021diffusion, ho2022classifier}, model selection~\citep{rombach2022high, ho2022cascaded, peebles2023scalable}, novel DDPM training paradigms~\citep{fan2023optimizing, black2023training} and related classes of generative modeling such as score/flow-matching~\citep{song2020score, lipman2023flowmatchinggenerativemodeling}.

\paragraph{Alignment of Diffusion Paths} Training deep learning models, such as a DDPM, is often done in two stages. In the first stage, the foundation model is optimized, where a general underlying data distribution is learned by training on large-scale datasets. In the second stage, this foundation model is fine-tuned to align to a reference distribution (e.g., towards downstream tasks, human preference, or stitching with multiple distributions) through supervised fine-tuning (SFT)~\citep{ziegler2019fine}, RL with AI/human feedback (RLAIF/RLHF)~\citep{bai2022training, ouyang2022training}, preference optimization (DPO)~\citep{rafailov2024direct, wallace2024diffusion, xiao2024comprehensivesurveydirectpreference}, or reward gradients~\citep{prabhudesai2023aligning, clark2023directly}. In this work, we apply and modify these techniques toward DMCs to better align using a scalar feedback signal that defines the reward distribution for sequential decision-making tasks.

\paragraph{Diffusion Models for Control (DMC)} Like many generative modeling algorithms, the role of DDPMs in decision-making applications has spanned from diffusion-based planning~\citep{janner2022planning} to policy learning~\citep{ren2024diffusion}, with significant successes and advancements in online and more prevalently in offline RL settings. Specifically, its utilization has been quite diverse, taking various forms such as diffusion-based policies, value functions, and world models, and these models can either indirectly support the training of RL policies~\citep{rigter2023world} or be more of an active component of the decision-making loop~\citep{wang2022diffusion, chen2022offline}. In this work, we extend these prior efforts, specifically Diffuser~\citep{janner2022planning}, Diffusion Policy (DP)~\citep{chi2023diffusion}, SynthER~\citep{lu2024synthetic} and some of their following works~\citep{ajay2023is, dong2024diffuserlite} to act as our foundation to build upon with our alignment efforts.

\section{Reward Alignment of DMC} \label{section: reward alignment}
We define a general equation that defines the reward alignment optimization problem, in which we align our diffusion model $\pi_\theta$ to generate samples $z_0$ that maximize a reward function.
\begin{equation}\label{equation: reward alignment}
    \mathcal{L}_{\text{align}}(\theta) := \underbrace{\mathbb{E}_{z_0 \sim \pi_\theta}[r(z_0)]}_{\text{Reward Maximization}}  \text{ s.t. } \underbrace{\text{div}(\pi_\theta, X) \leq \delta}_{\text{Divergence Constraint}}
\end{equation}
The constraint subjects the optimization to ensure the generated samples remain within a plausible region of the target data distribution $X$, or the distribution learned by the foundation DDPM $\pi_{\text{fdt}}$.

\begin{figure}
    \centering
    \includegraphics[width=\linewidth]{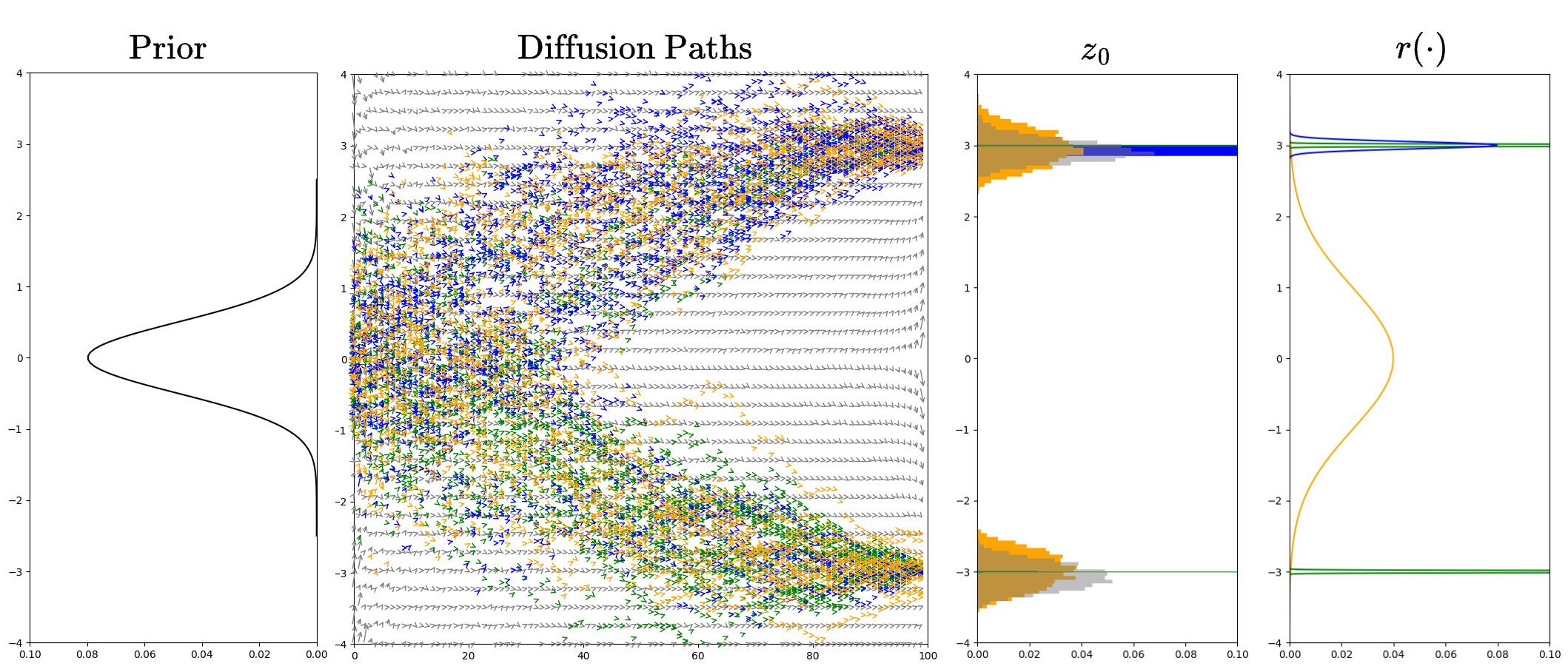}
    \caption{A visualization of the effects of alignment to the diffusion process. Given a foundation DDPM, whose outputs and diffusion field over the entire state space are shown in \textcolor{gray}{gray}, we align its diffusion process to maximize three separate reward functions, shown in \textcolor{orange}{orange}, \textcolor{green}{green} and \textcolor{blue}{blue}.}
    \label{fig:example}
\end{figure}

We illustrate this objective in Figure \ref{fig:example}, where a foundation DDPM is fine-tuned to maximize a given reward function on a toy task. In this example, the fine-tuned diffusion paths largely adhere to the foundation model, with minor yet impactful adjustments towards maximizing their respective reward functions, exceeding $\times3.21$, $\times7.09$, and $\times4.18$ improvements in \textcolor{orange}{orange}, \textcolor{green}{green} and \textcolor{blue}{blue} respectively within $32$ generated samples while maintaining $\text{Pr}(z_0|X) \gg 0$. Specifically, with the \textcolor{orange}{orange} reward function, despite having the potential to converge towards out-of-distribution (OOD) regions for maximal rewards, the constraints are in place to adequately address this issue in a desired manner. Hence, to achieve this alignment, we study and extend four approaches -- RL, DPO, SFT, and cascading -- to optimize the objective in Equation \ref{equation: reward alignment}.

\begin{figure}
\centering
\subfloat[Planning-based DMC]{\label{fig:diffuser}
\centering
\includegraphics[width=0.45\linewidth]{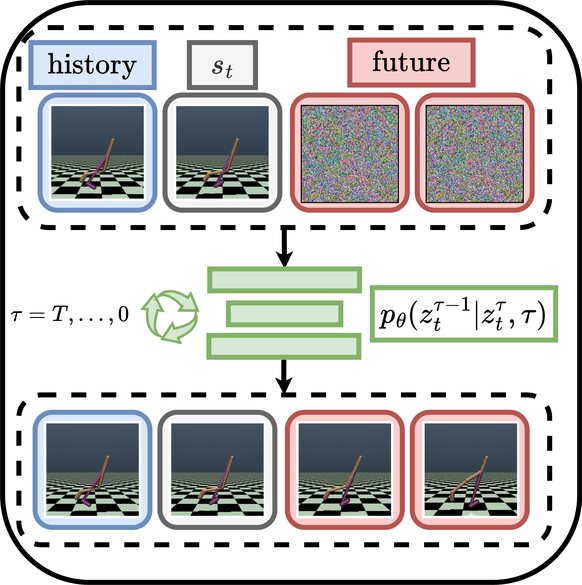}
}
%no space
\hfill
\subfloat[Policy-based DMC]{\label{fig:dp}
\centering

\includegraphics[width=0.45\linewidth]{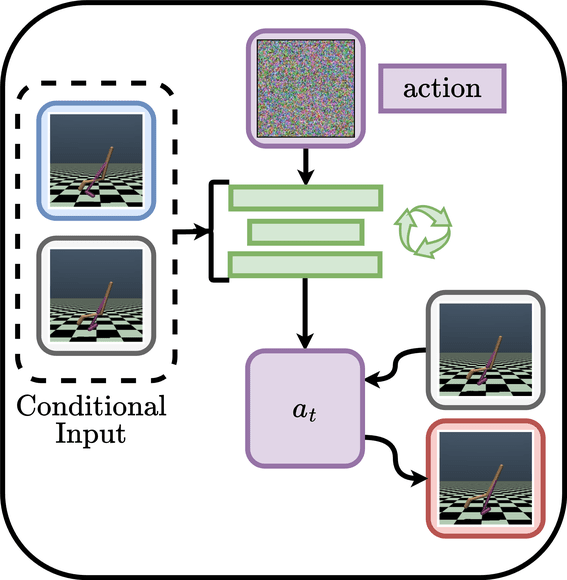}
}

\caption{A visual illustration of the two DMC frameworks on Walker2D task, where the planning-based DMC is forecasting future states and the policy-based DMC generates actions directly.}
\label{fig:dmc}
\end{figure}

An important distinction between DDPM and DMC is that, at denoising step $t$, DMC specifies that $z_t$ is either a $N-$step state-action trajectory $\{s_0, a_0, s_1 \dots s_N\}^t$ (e.g. Diffuser and DD), or actions $a^t_{t'}$ (e.g. DP) with a conditional information regarding the state $s_{t'}$ as shown in Figure \ref{fig:dmc}. Hence, our reward function $r(z_0)$ can either represent the return of the state-action trajectory or the Q-value of each state-action pair for greater granularity as a way to take advantage of the temporal aspect of the latent representation.

\subsection{Reward Alignment with Reinforcement Learning}
A natural approach to reward maximization is to extend traditional RL optimization to align the denoising process towards sample generation with higher returns. To do so, we view the DMC as a denoising policy
\begin{equation}
    \pi_\theta(\mu_t|z_t, c) = p_\theta(z_t | z_{t-1}, c)
\end{equation}
with a condition input $c$, treating the denoising process as a multi-step MDP~\citep{black2023training}. We build upon two classic RL algorithms that directly optimize Equation \ref{equation: reward alignment}: REINFORCE~\citep{williams1992simple} and Q-value policy gradient (QV-PG)~\citep{silver2014deterministic, haarnoja2018softactorcriticoffpolicymaximum}.

\begin{equation}\label{equation: reinfroce}
    \mathcal{L}_{\text{REINFORCE}}(\theta) = \mathbb{E}[\sum_{t=0}^T \kappa_t sg(\log \pi_\theta(\mu_t|z_t), K)\, r(z_0)]
\end{equation}
\begin{equation}\label{equation: qv-pg}
    \mathcal{L}_{\text{QV-PG}}(\theta) = \mathbb{E}[\kappa_t r( sg(\pi_\theta(\mu_{0:T}|z_{0:T}), K))]
\end{equation}
To account for the sequential nature of the denoising process, we consider two important training concerns: gradient truncation and denoising credit assignment.
\begin{itemize}
    \item Gradient truncation clips the gradient such its calculation only propagates through the last $K$ steps~\citep{clark2023directly, prabhudesai2023aligning}, where $K$ can be dynamically set during training. This truncation implements truncated backpropagation through time/denoising steps, facilitating a more memory-efficient and stable optimization. In practice, we utilize the stop-gradient function $sg(\cdot, K)$, which truncates the gradient computation to the last $K$ denoising steps. 
    \item As each sampling step contributed differently towards reaching $z_0$, we consider the concept of credit assignment over the denoising steps. Empirically, we define a sequence of hyperparameters $\{\kappa_0, \kappa_1, \dots\}$ that scale the objective along the denoising steps. In this work, we consider two options, where $\kappa$ can be a decreasing sequence or $\kappa_t$ can be the similarity measure between $(z_t,z_{t-1})$ and $(z_T,z_0)$, a computation that is done retrospectively.
\end{itemize}

To prevent large divergence from the reference policy/foundation DMC and adhere to the constraint in Equation \ref{equation: reward alignment}, we can use trust regions with clipping, i.e., proximal policy optimization (PPO)~\citep{schulman2017proximal}, similar to the procedure introduced with DDPO~\citep{black2023training}, or add a KL regularization term. We found both REINFORCE/PPO and QV-PG approaches to be competitive, to which their comparison seems moot, and the one that performed better is kept during evaluation.

\begin{table*}
    \centering
    \caption{\textbf{A Summary of Alignment Method Comparison:} Reward terms and divergence control mechanisms.}
    \label{tab:methods}
    \begin{tabular}{l|l|l}
        \toprule
        \textbf{Method} & \textbf{Reward Maximization Term} & \textbf{Divergence Control} \\
        \midrule
        RL & Policy gradient (Eqs.~\ref{equation: reinfroce},~\ref{equation: qv-pg}) & KL regularization/PPO clipping \\
        DPO & Preference ranking via \(F_w - F_l\) & Regularization via \(\epsilon_{\text{ref}}\) \\
        SFT & Fine-tuning on high reward samples & Likelihood threshold using \(\pi_{\text{ref}}\) \\
        Cascading & Conditional generation \(p(x^r_0 | x^c_0)\) & Warm-starting + adherence to reference policy. \\
        \bottomrule
    \end{tabular}
\end{table*}

\subsection{Reward Alignment with Preference Optimization}
Direct preference optimization (DPO)~\citep{rafailov2024direct} offers an alternative approach to align DMC by utilizing a dataset of ranked pairs $x^l_0 \prec x^w_0$ of losing and winning samples and maximizing the likelihoods of the DMC's sample generation following the preference operator $\prec$. This preference operation is defined in this work as follows:
\begin{equation}
    x^l_0 \prec x^w_0 \implies \mathbb{E}[r(x^l_0)] < \mathbb{E}[r(x^w_0)]
\end{equation}
Following DiffusionDPO adaptation~\citep{wallace2024diffusion}, the DPO objective $\mathcal{L}_{\text{DPO}}(\theta)$ is simplified and can be expressed as follows:
\begin{equation}
    \mathcal{L}_{\text{DPO}}(\theta)
    = -\mathbb{E}[
    \log\sigma (-\gamma \omega(\lambda_t)(F_w(\theta) - F_l(\theta)))]
    \label{eq:loss-dpo-1}
\end{equation}
where $\sigma(\cdot)$ is the sigmoid function, $\lambda = \alpha_t^2 / \sigma_t^2$ is the signal-to-noise ratio, $\omega(\lambda_t)$ is a weighting function and
\begin{equation}
    F_*(\theta) = \| \epsilon^* - \epsilon_\theta(z_{t}^*,t)\|^2_2 - \|\epsilon^* - \epsilon_\text{ref}(z_{t}^*,t)\|^2_2
\end{equation}

A functional representation of the preference operator can be expressed either heuristically or, more generally, as a trajectory-wise value function~\citep{zhang2024flow}.

\subsection{Reward Alignment with Supervised Fine-Tuning}
Similar to DPO, we can instead fine-tune the foundation DMC only using the winning samples, or samples with the highest rewards, by updating along $\mathcal{L}_{\text{DDPM}}(\theta)$ on these samples. Therefore, we apply traditional diffusion updates to shift and concentrate the generative distribution towards higher rewards under a stable optimization scheme, i.e., supervised learning.

This process can be performed iteratively, using either the foundation or the fine-tuned DMC as a data synthesizer to generate samples of increasingly higher returns. However, if we use the fine-tuned DMC, we face potential instability and OOD issues. In this work, we prevent this using a threshold constraint that prevents samples with lower likelihoods of the foundation DMC from being used for an SFT update.

\subsection{Up-sampling Reward with Cascading Diffusion}
Cascading DDPM~\citep{ho2022cascaded} is an up-sampling technique used for super-resolution, where lower resolution samples are processed as the conditional inputs to sample higher resolution samples. This concept can be extended to control applications, where a DMC that is conditioned on $x^\text{cond}_0$ is trained to generate a sample $x^r_0$ that yields greater rewards, hence $r(x^\text{cond}_0) < r(x^r_0)$.

In prior efforts with similar motivations~\citep{zhang2024flow}, the objective followed classifier-free guidance (CFG), training with the conditional information at some probability $p \in [0,1]$. Since, in our case, the cascading DMC is connected to a foundation DMC, it does not necessarily have to emphasize the generation of the target distribution; we set $p = 1$. Transferring the trained weights from a foundation DMC as a warm-start proved beneficial and was likely sufficient to understand the target distribution.

An important consideration for training cascading DMC is determining the pairs  $(x^\text{cond}_0, x^r_0)$ of samples to condition and reference during training, where their alignment may be mismatched, especially for multi-step samples. To address this, when curating $(x^\text{cond}_0, x^r_0)$, we promote consistency between the pairs. This consistency can be achieved by in-painting a contiguous subset of the sample or sharing the latent state $z_t$ at a denoising step $t$. These approaches force some level of relations between the two samples, such that the cascading DMC avoids up-sampling towards random samples during inference.

To further boost the up-sampling capability of the cascading DMC, we can extend the other alignment approaches to greaten the difference between $r(x^\text{cond}_0), r(x^r_0)$ with proper modifications. Our exploration of this study is limited, and we encourage future works to research this topic further.

\subsection{An Unification of Reward Alignment Approaches}
The four alignment methods discussed above---RL, DPO, SFT, and cascading---share a unified goal of solving the constrained optimization problem defined by Equation~\ref{equation: reward alignment}. While each approach is suited for different use cases, with their unique advantages and limitations, we find that they can be applied together. As the cost of increasing model alignment efforts is modest relative to training the reference policy \citep{ouyang2022traininglanguagemodelsfollow}, we propose a sequential optimization scheme, where each alignment approach (i.e., RL, DPO, and SFT) is used individually, and repeatedly until convergence. This iterative sequential scheme is practiced to avoid potential catastrophic learning variances of a naive multi-objective optimization approach. We find that after $2-3$ passes of this sequence, there are diminishing returns in the reward maximization efforts. Again, the cascading module is trained separately and appended onto the fine-tuned model during inference. Similarly, we experimented with performing multiple passes through the cascading module but found high variances in the up-sampling after $2$ passes.

\subsection{Design and Training Choices}
\paragraph{Diffusion Sampling and Process} We apply the proposed alignment methods to widespread advancements to the classical DDPM formulation. For instance, accelerated sampling methods such as DDIM \cite{ho2020denoisingdiffusionprobabilisticmodels} and other representations of the iterative diffusion-like posterior, such as score-matching (SM) \cite{song2020denoising} and flow-matching (FM) \cite{lipman2023flowmatchinggenerativemodeling}, are also studied in this work under the Diffuser, DD and DP framework utilizing these alignment approaches.

\paragraph{In-painting} The in-painting ability of DDPMs refers to reconstructing missing regions within the data based on the context provided. In our alignment efforts, we rely on this capability to bootstrap training samples to enforce consistency and avoid undesired changes to the sample. To ensure that our model can in-paint effectively, we trained the foundation DMC directly on the in-painting task, where we utilize causal masks along the temporal axis following the Repaint procedure for conditioning \cite{lugmayr2022repaint}. This in-painting procedure is followed during inference for planning-based DMC. In our experiments, we found that including this masked training improved not only the DMC's ability to in-paint but also the overall performance of the DMC.

\paragraph{Context Window and Recurrence} The context window for the DMC defines the number of time steps contained in input for planners and conditional input for the policy. Under specific model architectures, such as a temporal convolution backbone, this window does not face significant computing constraints. However, we opt for a sliding window approach to handle this constraint for backbones that scale poorly along this dimension and for tasks with longer time horizons. We found the introduction of a recurrent state appended to the latent state to cause instability in some of the tasks, and even in certain tasks with less complex observations with greater partial observability, we found minor but negligible improvements.

\paragraph{Evaluation Function}
For all alignment methods, we use a separate Q-value function trained under implicit Q-learning in an offline RL fashion to evaluate the samples generated and calculate the learning signal. The Q-value can also be used during action-selection.

\paragraph{Online Fine-tuning}
We additionally study the inclusion of online fine-tuning, where we inject samples from online interactions during the alignment training to compare and analyze the changes in performance.

\paragraph{PEFT} For all alignment processes, we use LORA \cite{hu2021lora}, a parameter-efficient fine-tuning approach that introduces a separate set of low-rank weights on linear layers that are trained during fine-tuning. Separate LORA weights were not introduced between different alignment methods.

\section{Experiments}

\begin{table*}
\centering
\small
% \vspace{-10pt}
\caption{\small{\textbf{Evaluation Results of Planning-based DMC.} The performance of various sampling approaches for planning-based DMC. Results correspond to the average return over $64$ episode seeds with $16$ roll-outs per action step, where the arrow indicates the results after our proposed alignment fine-tuning process.}}
\label{tab:planningdmc}
\scalebox{0.7}{
\setlength{\tabcolsep}{14pt}
\begin{tabular}{llcccccc}
\toprule
\textbf{Dataset} & \textbf{Environment} & \multicolumn{4}{c}{\textbf{Diffuser}} & \textbf{DD}  \\ \cmidrule(l){3-7} 
& & DDPM & DDIM & SM & FM & DDPM & \\ 
\midrule
\multirow{3}{*}{Random} & HalfCheetah & $8.0 \shortrightarrow 12.7$ & $5.5 \shortrightarrow 11.3$ & $5.4 \shortrightarrow 11.5$ & $10.2 \shortrightarrow 15.9$ & $8.4 \shortrightarrow 13.5$ \\
 & Hopper & $0.5 \shortrightarrow 4.2$ & $0.4 \shortrightarrow 1.1$ & $0.2 \shortrightarrow 3.0$ & $0.3 \shortrightarrow 3.9$ & $0.5 \shortrightarrow 1.8$ \\
 & Walker2d & $3.2 \shortrightarrow 9.1$ & $4.4 \shortrightarrow 10.1$ & $3.6 \shortrightarrow 5.9$ & $3.4 \shortrightarrow 7.2$ & $3.1 \shortrightarrow 6.9$ \\ \midrule
 
\multirow{3}{*}{Medium-Expert} & HalfCheetah & $85.2 \shortrightarrow 97.8$ & $77.8 \shortrightarrow 94.1$ & $74.3 \shortrightarrow 94.3$ & $90.0 \shortrightarrow 102.1$ & $86.2 \shortrightarrow 93.5$ \\
 & Hopper & $52.5 \shortrightarrow 107.2$ & $50.2 \shortrightarrow 100.1$ & $66.2 \shortrightarrow 111.2$ & $54.8 \shortrightarrow 113.2$ & $53.1 \shortrightarrow 110.9$ \\
 & Walker2d & $76.5 \shortrightarrow 102.4$ & $48.4 \shortrightarrow 108.3$ & $65.1 \shortrightarrow 109.0$ & $49.1 \shortrightarrow 99.4$ & $67.1 \shortrightarrow 101.1$ \\ \midrule
 
\multirow{3}{*}{Medium} & HalfCheetah & $30.6 \shortrightarrow 48.3$ & $23.3 \shortrightarrow 50.6$ & $32.1 \shortrightarrow 54.2$ & $35.1 \shortrightarrow 50.2$ & $29.0 \shortrightarrow 44.5$ \\
 & Hopper & $52.9 \shortrightarrow 89.5$ & $61.2 \shortrightarrow 93.5$ & $49.0 \shortrightarrow 82.0$ & $44.2 \shortrightarrow 96.3$ & $37.5 \shortrightarrow 90.2$ \\
 & Walker2d & $35.3 \shortrightarrow 76.6$ & $29.6 \shortrightarrow 61.8$ & $36.1 \shortrightarrow 78.0$ & $33.4 \shortrightarrow 89.2$ & $29.1 \shortrightarrow 50.4$ \\ \midrule
 
\multirow{3}{*}{Medium-Replay} & HalfCheetah & $28.6 \shortrightarrow 36.4$ & $22.9 \shortrightarrow 34.1$ & $14.1 \shortrightarrow 26.5$ & $26.8 \shortrightarrow 37.9$ & $25.7 \shortrightarrow 36.4$ \\
& Hopper & $54.1 \shortrightarrow 91.8$ & $49.2 \shortrightarrow 94.8$ & $55.3 \shortrightarrow 77.9$ & $58.5 \shortrightarrow 89.4$ & $45.8 \shortrightarrow 70.5$ \\
 & Walker2d & $26.0 \shortrightarrow 58.6$ & $45.6 \shortrightarrow 67.5$ & $46.2 \shortrightarrow 59.1$ & $50.5 \shortrightarrow 62.7$ & $41.1 \shortrightarrow 52.4$ \\ \midrule
Mixed & Franka-Kitchen & $41.5 \shortrightarrow 56.5$ & $52.0 \shortrightarrow 55.9$ & $32.5 \shortrightarrow 54.2$ & $45.7 \shortrightarrow 52.0$ & $44.5 \shortrightarrow 60.3$ \\
Partial & Franka-Kitchen & $38.0 \shortrightarrow 40.1$ & $33.5 \shortrightarrow 34.2$ & $36.1 \shortrightarrow 41.2$ & $36.3 \shortrightarrow 50.1$ & $32.7 \shortrightarrow 44.9$ \\ \midrule
\multirow{2}{*}{Play} & Antmaze-Medium & $0.2 \shortrightarrow 46.7$ & $0.0 \shortrightarrow 45.3$ & $0.9 \shortrightarrow 84.0$ & $3.2 \shortrightarrow 80.8$ & $0.0 \shortrightarrow 32.1$ \\

& Antmaze-Large & $0.0 \shortrightarrow 57.3$ & $0.0 \shortrightarrow 60.2$ & $3.5 \shortrightarrow 76.1$ & $0.7 \shortrightarrow 54.9$ & $0.0 \shortrightarrow 85.1$ \\
 
\multirow{2}{*}{Diverse} & Antmaze-Medium & $0.8 \shortrightarrow 42.0$ & $4.0 \shortrightarrow 75.3$ & $10.1 \shortrightarrow 62.3$ & $11.2 \shortrightarrow 92.5$ & $2.4 \shortrightarrow 63.2$ \\

& Antmaze-Large & $0.0 \shortrightarrow 47.3$ & $0.0 \shortrightarrow 50.6$ & $8.9 \shortrightarrow 50.3$ & $0.0 \shortrightarrow 70.1$ & $0.8 \shortrightarrow 41.3$ \\ \midrule

\multirow{4}{*}{MetaWorld} & Button-Press & $0.0 \shortrightarrow 0.7$ & $0.1 \shortrightarrow 1.0$ & $0.0 \shortrightarrow 0.5$ & $0.0 \shortrightarrow 0.7$ & $0.1 \shortrightarrow 0.5$ \\
& Drawer-Open& $0.2 \shortrightarrow 0.9$ & $0.1 \shortrightarrow 0.9$ & $0.2 \shortrightarrow 0.7$ & $0.4 \shortrightarrow 0.6$ & $0.3 \shortrightarrow 0.9$ \\
& Sweep-Into & $0.3 \shortrightarrow 0.7$ & $0.0 \shortrightarrow 0.8$ & $0.5 \shortrightarrow 0.8$ & $0.3 \shortrightarrow 0.9$ & $0.2 \shortrightarrow 1.0$ \\
& Plate-Slide & $0.2 \shortrightarrow 0.6$ & $0.1 \shortrightarrow 0.4$ & $0.2 \shortrightarrow 0.7$ & $0.4 \shortrightarrow 0.7$ & $0.1 \shortrightarrow 0.6$ \\
\midrule

\multirow{3}{*}{Nav1D} & Partial & $2.1 \shortrightarrow 4.3$ & $1.9 \shortrightarrow 3.3$ & $2.5 \shortrightarrow 4.1$ & $2.3 \shortrightarrow 5.2$ & $2.0 \shortrightarrow 3.9$ \\
& Full& $3.2 \shortrightarrow 6.9$ & $4.3 \shortrightarrow 7.3$ & $2.9 \shortrightarrow 6.9$ & $3.9 \shortrightarrow 6.8$ & $3.3 \shortrightarrow 6.9$ \\
& Expert & $6.2 \shortrightarrow 7.3$ & $5.9 \shortrightarrow 7.2$ & $5.9 \shortrightarrow 7.2$ & $6.4 \shortrightarrow 7.1$ & $6.2 \shortrightarrow 7.2$ 
 \\ \bottomrule
\end{tabular}}
\end{table*}
\subsection{Tasks} We evaluate our alignment approaches on various offline RL benchmarks. We use D4RL and MetaWorld for planning-based DMC; for policy-based DMC, we use Robomimic, Relay-Kitchen, and PushT. There is a standardized and publicly available offline RL dataset for many of these tasks, and we direct readers toward their respective original works for more details.
\begin{figure}
    \centering
    \includegraphics[width=0.7\linewidth]{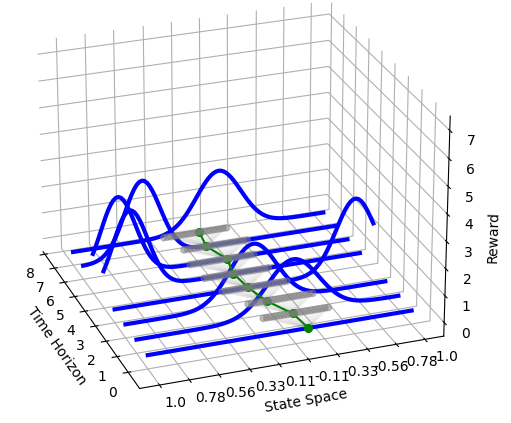}
    \caption{Visualization of the Nav1D task and an example of an agent's trajectory, shown in green, and the actions space is shown in gray. The reward distributions at every time step are shown in blue.}
    \label{fig:nav1d}
\end{figure}
\paragraph{Nav1D} We also introduce simple 1D navigation (Nav1D) task to our testbed, shown in Figure \ref{fig:nav1d}, where the agent start at a random position $s_t\in [-1,1]$ and must move left or right. The agent must select actions under a continuous yet restricted action space with a deterministic transition dynamics \begin{equation}
    \mathcal{T}(s_t,a_t) = \mathbb{I}(\text{clip}(s_t + \gamma a_t, -1, 1))
\end{equation}
where $\gamma$ is the step size. The objective is to maximize the rewards defined by a sequence of independent normal distribution $\{N_t\}_{t=0}^H$, where 
\begin{equation}
    r(s_t,a_t) \propto \text{Pr}(T(s_t,a_t) = N_{t+1})
\end{equation}
over a fixed time horizon $H$. The reward distributions are independent from one another and stationary upon repeated trials. The agent's observation space consists of its current location and additional information that gives the agent a better time-dependent context. 

We create three datasets for Nav1D that provide varying state coverages—partial, full, and expert—where both partial and full datasets follow a random policy, and the expert dataset contains trajectories that follow an expert policy.

\paragraph{Coherency Score} We introduce a coherency score $C$ to evaluate the sample quality of planning-based DMC,
\begin{equation}
    C = \mathbb{E}_{(s_t, a_t, s_{t+1})\sim \pi_\theta}[\text{Pr}(s_{t+1} = \mathcal{T}(s_t,a_t)]
\end{equation}
and of policy-based solutions,
\begin{equation}
    C = \mathbb{E}_{a_t\sim \pi_\theta}[\text{Pr}(a_t = \pi_\beta(a_t|s_t)]
\end{equation}
The coherency score measures how well the planning methods capture the dynamics model distribution, whereas, for the policy-based DMC, it measures how well the DMC models the behavior policy $\pi_\beta$. 

\subsection{Implementation Details} The basis of our implementation is built upon CleanDiffuser \cite{cleandiffuser}, with adjustments made for the alignment methods. A central difference in our implementation of the DMC is the lack of a return conditional input, as it is generally unknown what return is desired. Hence, for action selection, we use the learned Q-Value function to greedily select actions, similar to IDQL \cite{hansen2023idql}. For DP, we also experiment with expectation-based action selection, where the average over all generated action samples is taken. The model architecture for each algorithm we studied follows the original work, similar to CleanDiffuser. All experiments presented were executed on $3$ Nvidia RTX A6000 and Intel Xeon Silver 4214R @ 2.4GHz.

\subsection{Results For Planning-based DMC}
We summarize our finding for planning-based DMC in Table \ref{tab:planningdmc}, showing the notable improvements from our proposed alignment process, regardless of the sampling approaches such as DDPM, DDIM, SM, FM, and DD frameworks, with decision-making that uses a small number of roll-outs during its planning. Specifically, for each task and algorithm, we first trained the reference DMC and finetuned it using varying passes and iterations through \{RL, DPO, SFT\} alignments.

\begin{figure}
    \centering
    \includegraphics[width=\linewidth]{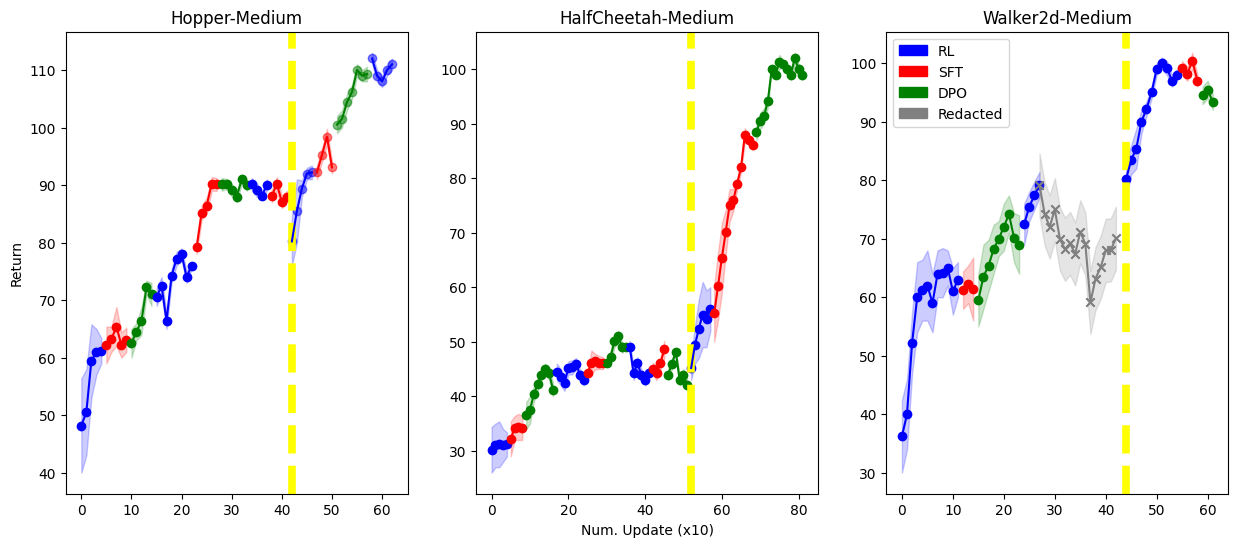}
    \caption{Alignment Learning Curve for Diffuser (DDPM) on the D4RL Medium Datasets, where the average and $\pm1$ standard deviation is shown over 64 episode seeds. The yellow vertical line indicates when the online fine-tuning is introduced.}
    \label{fig:curve}
\end{figure}

We closely examined the learning for these training to ensure no training collapses occurred, with model check-pointing and lowered learning rates. This phenomenon includes sudden drops in return and instabilities/spikes in the training objectives. We did observe some instances of this, especially during the RL finetuning, which we redacted. We trained the cascading DDPM separately, applying the same alignment process, and attached it for the final evaluation. In general, we found that even with multiple rounds, the return did improve, as shown in Figure \ref{fig:curve}.

\begin{figure}
    \centering
    \includegraphics[width=0.9\linewidth]{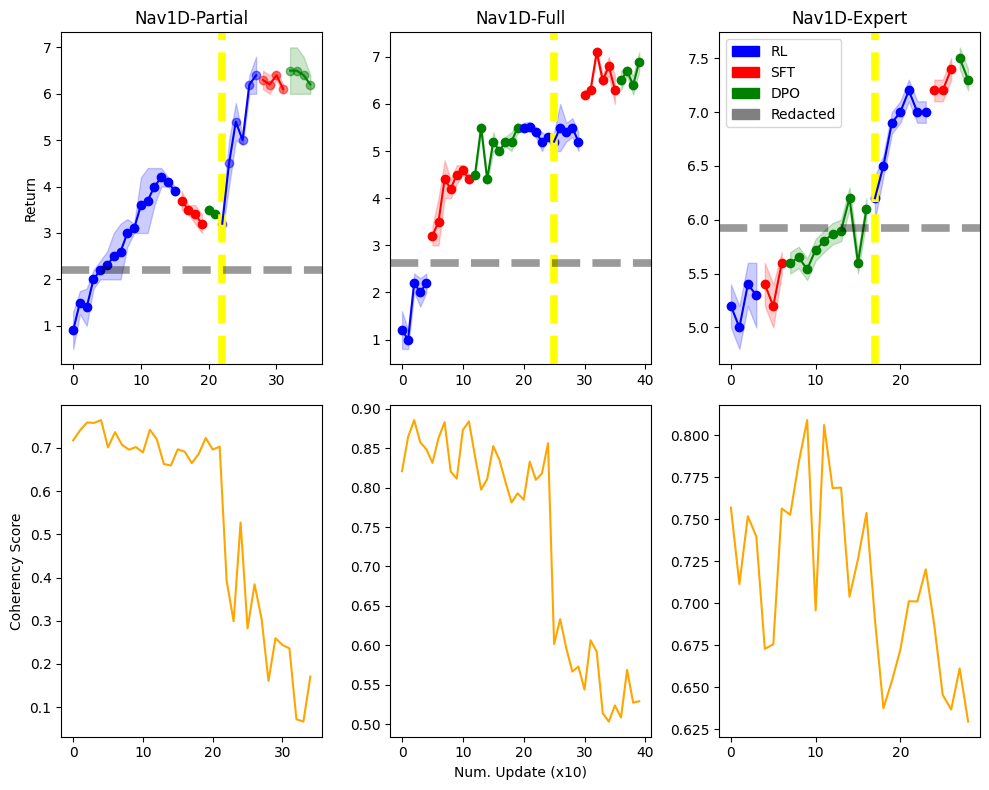}
    \caption{Alignment Learning Curve and Coherency Score for DP (Expectation) on the Nav1D Datasets, where the average and $\pm1$ standard deviation is shown over 32 episode seeds. The yellow vertical line indicates when the online fine-tuning is introduced and the black horizontal line represents the average return achieved in the respective dataset.}
    \label{fig:curve2}
\end{figure}

An important trend we noticed as a result of the alignment is that, on average, over all tasks and algorithms, the average standard deviation over the return on all trials significantly decreased, from $38.1\% \rightarrow 5.3\%$. This result empirically indicates the distributional shift that occurs, which narrows the diffusion process towards higher returns, as the average return improved by $+43.0\%$. Using only the D4RL tasks, we introduced an entropy regularization term to counter this distributional narrowing as much as possible. However, this resulted in the average return improvement to decrease from $32.2\%\rightarrow20.3\%$.

Additionally, with the alignment process under the Nav1D task, we found the coherency score remained relatively stable with repeated passes and iterations. We monitored this value during training to ensure stable learning, tuning the divergence control hyperparameters as needed. However, we did notice large decreases in the coherency score when we performed multiple passes through the cascading DDPM.

\begin{table}
\centering
\small
% \vspace{-10pt}
\caption{\small{\textbf{Evaluation Results of Policy-based DMC.} The performance of two action-selection methods (expectation vs IDQL policy extraction) for policy-based DMC. Results correspond to the average return over $16$ episode seeds on $8$ action candidates, where the arrow indicates the results after our proposed alignment fine-tuning process.}}
\label{tab:policydmc}
\scalebox{0.7}{
\setlength{\tabcolsep}{14pt}
\begin{tabular}{llcc}
\toprule
\textbf{Dataset} & \textbf{Environment} & \multicolumn{2}{c}{\textbf{DP}}  \\ \cmidrule(l){3-4} 
& & Expectation & IDQL \\ 
\midrule
\multirow{10}{*}{Low-Dim} & pusht & $0.42 \shortrightarrow 1.0$ & $0.71 \shortrightarrow 1.0$  \\
&  relay-kitchen & $0.75 \shortrightarrow 1.0$ & $0.98 \shortrightarrow 1.0$  \\
&  lift-ph & $0.61 \shortrightarrow 1.0$ & $1.0 \shortrightarrow 1.0$  \\
&  lift-mh & $0.65 \shortrightarrow 0.98$ & $0.79 \shortrightarrow 1.0$  \\
&  can-ph & $0.51 \shortrightarrow 1.0$ & $0.92 \shortrightarrow 1.0$  \\
&  can-mh & $0.47 \shortrightarrow 0.95$ & $0.89 \shortrightarrow 1.0$  \\
&  square-ph & $0.32 \shortrightarrow 0.55$ & $0.60 \shortrightarrow 0.95$  \\
&  square-mh & $0.40 \shortrightarrow 0.67$ & $0.58 \shortrightarrow 0.93$  \\
&  transport-ph & $0.58 \shortrightarrow 1.0$ & $0.74 \shortrightarrow 1.0$  \\
&  transport-mh & $0.45 \shortrightarrow 0.91$ & $0.88 \shortrightarrow 1.0$  \\
\midrule

\multirow{9}{*}{Image} & pusht-image & $0.62 \shortrightarrow 1.0$ & $0.91 \shortrightarrow 1.0$  \\
&  lift-ph & $0.41 \shortrightarrow 1.0$ & $1.0 \shortrightarrow 1.0$  \\
&  lift-mh & $0.85 \shortrightarrow 1.0$ & $0.90 \shortrightarrow 1.0$  \\
&  can-ph & $0.51 \shortrightarrow 0.90$ & $0.71 \shortrightarrow 0.98$  \\
&  can-mh & $0.39 \shortrightarrow 0.98$ & $0.56 \shortrightarrow 0.99$  \\
&  square-ph & $0.21 \shortrightarrow 0.76$ & $0.67 \shortrightarrow 0.80$  \\
&  square-mh & $0.30 \shortrightarrow 0.45$ & $0.58 \shortrightarrow 0.89$  \\
&  transport-ph & $0.42 \shortrightarrow 0.77$ & $0.70 \shortrightarrow 0.79$  \\
&  transport-mh & $0.39 \shortrightarrow 0.55$ & $0.74 \shortrightarrow 0.78$\\
\midrule

\multirow{3}{*}{Nav1D} & partial & $1.10 \shortrightarrow 3.91$ & $2.69 \shortrightarrow 4.42$  \\
&  full & $1.32 \shortrightarrow 5.98$ & $4.61 \shortrightarrow 6.24$  \\
&  expert & $5.43 \shortrightarrow 5.80$ & $6.89 \shortrightarrow 7.06$  
 \\ \bottomrule
\end{tabular}}
\end{table}

Lastly, we experimented with online fine-tuning, enabling interactions with the environment during the alignment process to collect more data points. We also continued training the Q-value function on these samples. We evaluated this inclusion under D4RL benchmark tasks, and we observed a significantly more stable and efficient training stage with no training collapses and notable improvements in the return achieved.

\subsection{Results For Policy-based DMC}
In Table \ref{tab:policydmc}, we evaluate the performance of Diffusion Policy (DP) using two different action selection methods using the DiT architecture \cite{peebles2023scalable}. Similar to our approach for the planning-based DMC, we applied the alignment approach in sequential order and found multiple rounds still to have return improvements. We find overall improvements in return with our proposed alignment process, more so with the expectation action selection models. In some cases, we found the energy-based action selection approaches such as IDQL are sufficient to achieve perfect success rates, where we still observed that the fine-tuning did affect the narrowness of the posterior, measured by significantly lower entropy between the generated action candidates.

Regarding the coherency score in the Nav1D task, we found that the alignment process did shift the action distribution away from the behavioral policy, more notably in models trained on the partial and full datasets and upon online fine-tuning. This trend was not as starkly observed in the expert dataset, which retained a comparable coherency score from its reference counterpart.

\section{Conclusion}
In this work, we studied four alignment methods -- RL, DPO, SFT, and cascading -- for reward maximization in diffusion-based control models (DMC). While these alignment methods were applied independently in prior works, we proposed a unification of these fine-tuning processes using a sequential optimization scheme. Our findings were evaluated in both planning-based and policy-based DMC on various control tasks, and we established the empirical utility of using our proposed alignment approach. For future directions, we suggest finding methods of automating the sequential process, avoiding manual progression on the alignment sequence upon convergence or training failures.

\bibliography{example_paper}

\begin{thebibliography}{52}
\providecommand{\natexlab}[1]{#1}
\providecommand{\url}[1]{\texttt{#1}}
\expandafter\ifx\csname urlstyle\endcsname\relax
  \providecommand{\doi}[1]{doi: #1}\else
  \providecommand{\doi}{doi: \begingroup \urlstyle{rm}\Url}\fi

\bibitem[Ajay et~al.(2023)Ajay, Du, Gupta, Tenenbaum, Jaakkola, and Agrawal]{ajay2023is}
Ajay, A., Du, Y., Gupta, A., Tenenbaum, J.~B., Jaakkola, T.~S., and Agrawal, P.
\newblock Is conditional generative modeling all you need for decision making?
\newblock In \emph{The Eleventh International Conference on Learning Representations}, 2023.
\newblock URL \url{https://openreview.net/forum?id=sP1fo2K9DFG}.

\bibitem[Bai et~al.(2022)Bai, Jones, Ndousse, Askell, Chen, DasSarma, Drain, Fort, Ganguli, Henighan, et~al.]{bai2022training}
Bai, Y., Jones, A., Ndousse, K., Askell, A., Chen, A., DasSarma, N., Drain, D., Fort, S., Ganguli, D., Henighan, T., et~al.
\newblock Training a helpful and harmless assistant with reinforcement learning from human feedback.
\newblock \emph{arXiv preprint arXiv:2204.05862}, 2022.

\bibitem[Black et~al.(2023)Black, Janner, Du, Kostrikov, and Levine]{black2023training}
Black, K., Janner, M., Du, Y., Kostrikov, I., and Levine, S.
\newblock Training diffusion models with reinforcement learning.
\newblock \emph{arXiv preprint arXiv:2305.13301}, 2023.

\bibitem[Cao et~al.(2024)Cao, Tan, Gao, Xu, Chen, Heng, and Li]{cao2024survey}
Cao, H., Tan, C., Gao, Z., Xu, Y., Chen, G., Heng, P.-A., and Li, S.~Z.
\newblock A survey on generative diffusion models.
\newblock \emph{IEEE Transactions on Knowledge and Data Engineering}, 2024.

\bibitem[Chen et~al.(2022)Chen, Lu, Ying, Su, and Zhu]{chen2022offline}
Chen, H., Lu, C., Ying, C., Su, H., and Zhu, J.
\newblock Offline reinforcement learning via high-fidelity generative behavior modeling.
\newblock \emph{arXiv preprint arXiv:2209.14548}, 2022.

\bibitem[Chi et~al.(2023)Chi, Xu, Feng, Cousineau, Du, Burchfiel, Tedrake, and Song]{chi2023diffusion}
Chi, C., Xu, Z., Feng, S., Cousineau, E., Du, Y., Burchfiel, B., Tedrake, R., and Song, S.
\newblock Diffusion policy: Visuomotor policy learning via action diffusion.
\newblock \emph{The International Journal of Robotics Research}, pp.\  02783649241273668, 2023.

\bibitem[Clark et~al.(2023)Clark, Vicol, Swersky, and Fleet]{clark2023directly}
Clark, K., Vicol, P., Swersky, K., and Fleet, D.~J.
\newblock Directly fine-tuning diffusion models on differentiable rewards.
\newblock \emph{arXiv preprint arXiv:2309.17400}, 2023.

\bibitem[Dhariwal \& Nichol(2021)Dhariwal and Nichol]{dhariwal2021diffusion}
Dhariwal, P. and Nichol, A.
\newblock Diffusion models beat gans on image synthesis.
\newblock \emph{Advances in neural information processing systems}, 34:\penalty0 8780--8794, 2021.

\bibitem[Dong et~al.(2024{\natexlab{a}})Dong, Hao, Yuan, Ni, Wang, Li, and Zheng]{dong2024diffuserlite}
Dong, Z., Hao, J., Yuan, Y., Ni, F., Wang, Y., Li, P., and Zheng, Y.
\newblock Diffuserlite: Towards real-time diffusion planning.
\newblock \emph{arXiv preprint arXiv:2401.15443}, 2024{\natexlab{a}}.

\bibitem[Dong et~al.(2024{\natexlab{b}})Dong, Yuan, Hao, Ni, Ma, Li, and Zheng]{cleandiffuser}
Dong, Z., Yuan, Y., Hao, J., Ni, F., Ma, Y., Li, P., and Zheng, Y.
\newblock Cleandiffuser: An easy-to-use modularized library for diffusion models in decision making.
\newblock \emph{arXiv preprint arXiv:2406.09509}, 2024{\natexlab{b}}.
\newblock URL \url{https://arxiv.org/abs/2406.09509}.

\bibitem[Fan \& Lee(2023)Fan and Lee]{fan2023optimizing}
Fan, Y. and Lee, K.
\newblock Optimizing ddpm sampling with shortcut fine-tuning.
\newblock \emph{arXiv preprint arXiv:2301.13362}, 2023.

\bibitem[Florence et~al.(2022)Florence, Lynch, Zeng, Ramirez, Wahid, Downs, Wong, Lee, Mordatch, and Tompson]{florence2022implicit}
Florence, P., Lynch, C., Zeng, A., Ramirez, O.~A., Wahid, A., Downs, L., Wong, A., Lee, J., Mordatch, I., and Tompson, J.
\newblock Implicit behavioral cloning.
\newblock In \emph{Conference on Robot Learning}, pp.\  158--168. PMLR, 2022.

\bibitem[Fu et~al.(2020)Fu, Kumar, Nachum, Tucker, and Levine]{fu2020d4rl}
Fu, J., Kumar, A., Nachum, O., Tucker, G., and Levine, S.
\newblock D4rl: Datasets for deep data-driven reinforcement learning.
\newblock \emph{arXiv preprint arXiv:2004.07219}, 2020.

\bibitem[Gupta et~al.(2019)Gupta, Kumar, Lynch, Levine, and Hausman]{gupta2019relaypolicylearningsolving}
Gupta, A., Kumar, V., Lynch, C., Levine, S., and Hausman, K.
\newblock Relay policy learning: Solving long-horizon tasks via imitation and reinforcement learning, 2019.
\newblock URL \url{https://arxiv.org/abs/1910.11956}.

\bibitem[Haarnoja et~al.(2018)Haarnoja, Zhou, Abbeel, and Levine]{haarnoja2018softactorcriticoffpolicymaximum}
Haarnoja, T., Zhou, A., Abbeel, P., and Levine, S.
\newblock Soft actor-critic: Off-policy maximum entropy deep reinforcement learning with a stochastic actor, 2018.
\newblock URL \url{https://arxiv.org/abs/1801.01290}.

\bibitem[Hansen-Estruch et~al.(2023)Hansen-Estruch, Kostrikov, Janner, Kuba, and Levine]{hansen2023idql}
Hansen-Estruch, P., Kostrikov, I., Janner, M., Kuba, J.~G., and Levine, S.
\newblock Idql: Implicit q-learning as an actor-critic method with diffusion policies.
\newblock \emph{arXiv preprint arXiv:2304.10573}, 2023.

\bibitem[Ho \& Salimans(2022)Ho and Salimans]{ho2022classifier}
Ho, J. and Salimans, T.
\newblock Classifier-free diffusion guidance.
\newblock \emph{arXiv preprint arXiv:2207.12598}, 2022.

\bibitem[Ho et~al.(2020)Ho, Jain, and Abbeel]{ho2020denoisingdiffusionprobabilisticmodels}
Ho, J., Jain, A., and Abbeel, P.
\newblock Denoising diffusion probabilistic models, 2020.
\newblock URL \url{https://arxiv.org/abs/2006.11239}.

\bibitem[Ho et~al.(2022)Ho, Saharia, Chan, Fleet, Norouzi, and Salimans]{ho2022cascaded}
Ho, J., Saharia, C., Chan, W., Fleet, D.~J., Norouzi, M., and Salimans, T.
\newblock Cascaded diffusion models for high fidelity image generation.
\newblock \emph{Journal of Machine Learning Research}, 23\penalty0 (47):\penalty0 1--33, 2022.

\bibitem[Hu et~al.(2021)Hu, Shen, Wallis, Allen-Zhu, Li, Wang, Wang, and Chen]{hu2021lora}
Hu, E.~J., Shen, Y., Wallis, P., Allen-Zhu, Z., Li, Y., Wang, S., Wang, L., and Chen, W.
\newblock Lora: Low-rank adaptation of large language models.
\newblock \emph{arXiv preprint arXiv:2106.09685}, 2021.

\bibitem[Janner et~al.(2022)Janner, Du, Tenenbaum, and Levine]{janner2022planning}
Janner, M., Du, Y., Tenenbaum, J.~B., and Levine, S.
\newblock Planning with diffusion for flexible behavior synthesis.
\newblock \emph{arXiv preprint arXiv:2205.09991}, 2022.

\bibitem[Kingma et~al.(2021)Kingma, Salimans, Poole, and Ho]{kingma2021variational}
Kingma, D., Salimans, T., Poole, B., and Ho, J.
\newblock Variational diffusion models.
\newblock \emph{Advances in neural information processing systems}, 34:\penalty0 21696--21707, 2021.

\bibitem[Lipman et~al.(2023)Lipman, Chen, Ben-Hamu, Nickel, and Le]{lipman2023flowmatchinggenerativemodeling}
Lipman, Y., Chen, R. T.~Q., Ben-Hamu, H., Nickel, M., and Le, M.
\newblock Flow matching for generative modeling, 2023.
\newblock URL \url{https://arxiv.org/abs/2210.02747}.

\bibitem[Liu et~al.(2022)Liu, Gong, and Liu]{liu2022flow}
Liu, X., Gong, C., and Liu, Q.
\newblock Flow straight and fast: Learning to generate and transfer data with rectified flow.
\newblock \emph{arXiv preprint arXiv:2209.03003}, 2022.

\bibitem[Lu et~al.(2024)Lu, Ball, Teh, and Parker-Holder]{lu2024synthetic}
Lu, C., Ball, P., Teh, Y.~W., and Parker-Holder, J.
\newblock Synthetic experience replay.
\newblock \emph{Advances in Neural Information Processing Systems}, 36, 2024.

\bibitem[Lugmayr et~al.(2022)Lugmayr, Danelljan, Romero, Yu, Timofte, and Van~Gool]{lugmayr2022repaint}
Lugmayr, A., Danelljan, M., Romero, A., Yu, F., Timofte, R., and Van~Gool, L.
\newblock Repaint: Inpainting using denoising diffusion probabilistic models.
\newblock In \emph{Proceedings of the IEEE/CVF conference on computer vision and pattern recognition}, pp.\  11461--11471, 2022.

\bibitem[Mandlekar et~al.(2021)Mandlekar, Xu, Wong, Nasiriany, Wang, Kulkarni, Fei-Fei, Savarese, Zhu, and Mart{\'\i}n-Mart{\'\i}n]{mandlekar2021matters}
Mandlekar, A., Xu, D., Wong, J., Nasiriany, S., Wang, C., Kulkarni, R., Fei-Fei, L., Savarese, S., Zhu, Y., and Mart{\'\i}n-Mart{\'\i}n, R.
\newblock What matters in learning from offline human demonstrations for robot manipulation.
\newblock \emph{arXiv preprint arXiv:2108.03298}, 2021.

\bibitem[Nichol \& Dhariwal(2021)Nichol and Dhariwal]{nichol2021improved}
Nichol, A.~Q. and Dhariwal, P.
\newblock Improved denoising diffusion probabilistic models.
\newblock In \emph{International conference on machine learning}, pp.\  8162--8171. PMLR, 2021.

\bibitem[Ouyang et~al.(2022{\natexlab{a}})Ouyang, Wu, Jiang, Almeida, Wainwright, Mishkin, Zhang, Agarwal, Slama, Ray, et~al.]{ouyang2022training}
Ouyang, L., Wu, J., Jiang, X., Almeida, D., Wainwright, C., Mishkin, P., Zhang, C., Agarwal, S., Slama, K., Ray, A., et~al.
\newblock Training language models to follow instructions with human feedback.
\newblock \emph{Advances in neural information processing systems}, 35:\penalty0 27730--27744, 2022{\natexlab{a}}.

\bibitem[Ouyang et~al.(2022{\natexlab{b}})Ouyang, Wu, Jiang, Almeida, Wainwright, Mishkin, Zhang, Agarwal, Slama, Ray, Schulman, Hilton, Kelton, Miller, Simens, Askell, Welinder, Christiano, Leike, and Lowe]{ouyang2022traininglanguagemodelsfollow}
Ouyang, L., Wu, J., Jiang, X., Almeida, D., Wainwright, C.~L., Mishkin, P., Zhang, C., Agarwal, S., Slama, K., Ray, A., Schulman, J., Hilton, J., Kelton, F., Miller, L., Simens, M., Askell, A., Welinder, P., Christiano, P., Leike, J., and Lowe, R.
\newblock Training language models to follow instructions with human feedback, 2022{\natexlab{b}}.
\newblock URL \url{https://arxiv.org/abs/2203.02155}.

\bibitem[Peebles \& Xie(2023)Peebles and Xie]{peebles2023scalable}
Peebles, W. and Xie, S.
\newblock Scalable diffusion models with transformers.
\newblock In \emph{Proceedings of the IEEE/CVF International Conference on Computer Vision}, pp.\  4195--4205, 2023.

\bibitem[Prabhudesai et~al.(2023)Prabhudesai, Goyal, Pathak, and Fragkiadaki]{prabhudesai2023aligning}
Prabhudesai, M., Goyal, A., Pathak, D., and Fragkiadaki, K.
\newblock Aligning text-to-image diffusion models with reward backpropagation.
\newblock \emph{arXiv preprint arXiv:2310.03739}, 2023.

\bibitem[Rafailov et~al.(2024)Rafailov, Sharma, Mitchell, Manning, Ermon, and Finn]{rafailov2024direct}
Rafailov, R., Sharma, A., Mitchell, E., Manning, C.~D., Ermon, S., and Finn, C.
\newblock Direct preference optimization: Your language model is secretly a reward model.
\newblock \emph{Advances in Neural Information Processing Systems}, 36, 2024.

\bibitem[Ren et~al.(2024)Ren, Lidard, Ankile, Simeonov, Agrawal, Majumdar, Burchfiel, Dai, and Simchowitz]{ren2024diffusion}
Ren, A.~Z., Lidard, J., Ankile, L.~L., Simeonov, A., Agrawal, P., Majumdar, A., Burchfiel, B., Dai, H., and Simchowitz, M.
\newblock Diffusion policy policy optimization.
\newblock \emph{arXiv preprint arXiv:2409.00588}, 2024.

\bibitem[Rigter et~al.(2023)Rigter, Yamada, and Posner]{rigter2023world}
Rigter, M., Yamada, J., and Posner, I.
\newblock World models via policy-guided trajectory diffusion.
\newblock \emph{arXiv preprint arXiv:2312.08533}, 2023.

\bibitem[Rombach et~al.(2022)Rombach, Blattmann, Lorenz, Esser, and Ommer]{rombach2022high}
Rombach, R., Blattmann, A., Lorenz, D., Esser, P., and Ommer, B.
\newblock High-resolution image synthesis with latent diffusion models.
\newblock In \emph{Proceedings of the IEEE/CVF conference on computer vision and pattern recognition}, pp.\  10684--10695, 2022.

\bibitem[Salimans \& Ho(2022)Salimans and Ho]{salimans2022progressive}
Salimans, T. and Ho, J.
\newblock Progressive distillation for fast sampling of diffusion models.
\newblock \emph{arXiv preprint arXiv:2202.00512}, 2022.

\bibitem[Schulman et~al.(2017)Schulman, Wolski, Dhariwal, Radford, and Klimov]{schulman2017proximal}
Schulman, J., Wolski, F., Dhariwal, P., Radford, A., and Klimov, O.
\newblock Proximal policy optimization algorithms.
\newblock \emph{arXiv preprint arXiv:1707.06347}, 2017.

\bibitem[Silver et~al.(2014)Silver, Lever, Heess, Degris, Wierstra, and Riedmiller]{silver2014deterministic}
Silver, D., Lever, G., Heess, N., Degris, T., Wierstra, D., and Riedmiller, M.
\newblock Deterministic policy gradient algorithms.
\newblock In \emph{International conference on machine learning}, pp.\  387--395. Pmlr, 2014.

\bibitem[Sohl-Dickstein et~al.(2015)Sohl-Dickstein, Weiss, Maheswaranathan, and Ganguli]{sohl2015deep}
Sohl-Dickstein, J., Weiss, E., Maheswaranathan, N., and Ganguli, S.
\newblock Deep unsupervised learning using nonequilibrium thermodynamics.
\newblock In \emph{International conference on machine learning}, pp.\  2256--2265. PMLR, 2015.

\bibitem[Song et~al.(2020{\natexlab{a}})Song, Meng, and Ermon]{song2020denoising}
Song, J., Meng, C., and Ermon, S.
\newblock Denoising diffusion implicit models.
\newblock \emph{arXiv preprint arXiv:2010.02502}, 2020{\natexlab{a}}.

\bibitem[Song et~al.(2020{\natexlab{b}})Song, Sohl-Dickstein, Kingma, Kumar, Ermon, and Poole]{song2020score}
Song, Y., Sohl-Dickstein, J., Kingma, D.~P., Kumar, A., Ermon, S., and Poole, B.
\newblock Score-based generative modeling through stochastic differential equations.
\newblock \emph{arXiv preprint arXiv:2011.13456}, 2020{\natexlab{b}}.

\bibitem[Wallace et~al.(2024)Wallace, Dang, Rafailov, Zhou, Lou, Purushwalkam, Ermon, Xiong, Joty, and Naik]{wallace2024diffusion}
Wallace, B., Dang, M., Rafailov, R., Zhou, L., Lou, A., Purushwalkam, S., Ermon, S., Xiong, C., Joty, S., and Naik, N.
\newblock Diffusion model alignment using direct preference optimization.
\newblock In \emph{Proceedings of the IEEE/CVF Conference on Computer Vision and Pattern Recognition}, pp.\  8228--8238, 2024.

\bibitem[Wang et~al.(2024)Wang, Yang, Huang, Wang, and Li]{wang2024rectified}
Wang, F.-Y., Yang, L., Huang, Z., Wang, M., and Li, H.
\newblock Rectified diffusion: Straightness is not your need in rectified flow.
\newblock \emph{arXiv preprint arXiv:2410.07303}, 2024.

\bibitem[Wang et~al.(2022)Wang, Hunt, and Zhou]{wang2022diffusion}
Wang, Z., Hunt, J.~J., and Zhou, M.
\newblock Diffusion policies as an expressive policy class for offline reinforcement learning.
\newblock \emph{arXiv preprint arXiv:2208.06193}, 2022.

\bibitem[Williams(1992)]{williams1992simple}
Williams, R.~J.
\newblock Simple statistical gradient-following algorithms for connectionist reinforcement learning.
\newblock \emph{Machine learning}, 8:\penalty0 229--256, 1992.

\bibitem[Xiao et~al.(2024)Xiao, Wang, Gan, Zhao, He, Tuan, Chen, Jiang, Zhao, and Wu]{xiao2024comprehensivesurveydirectpreference}
Xiao, W., Wang, Z., Gan, L., Zhao, S., He, W., Tuan, L.~A., Chen, L., Jiang, H., Zhao, Z., and Wu, F.
\newblock A comprehensive survey of direct preference optimization: Datasets, theories, variants, and applications, 2024.
\newblock URL \url{https://arxiv.org/abs/2410.15595}.

\bibitem[Yang et~al.(2024)Yang, Zhang, Song, Hong, Xu, Zhao, Zhang, Cui, and Yang]{yang2024diffusionmodelscomprehensivesurvey}
Yang, L., Zhang, Z., Song, Y., Hong, S., Xu, R., Zhao, Y., Zhang, W., Cui, B., and Yang, M.-H.
\newblock Diffusion models: A comprehensive survey of methods and applications, 2024.
\newblock URL \url{https://arxiv.org/abs/2209.00796}.

\bibitem[Yu et~al.(2021)Yu, Quillen, He, Julian, Narayan, Shively, Bellathur, Hausman, Finn, and Levine]{yu2021metaworldbenchmarkevaluationmultitask}
Yu, T., Quillen, D., He, Z., Julian, R., Narayan, A., Shively, H., Bellathur, A., Hausman, K., Finn, C., and Levine, S.
\newblock Meta-world: A benchmark and evaluation for multi-task and meta reinforcement learning, 2021.
\newblock URL \url{https://arxiv.org/abs/1910.10897}.

\bibitem[Zhang et~al.(2024)Zhang, Sun, Ye, Liu, Zhang, and Yu]{zhang2024flow}
Zhang, Z., Sun, Y., Ye, J., Liu, T.-S., Zhang, J., and Yu, Y.
\newblock Flow to better: Offline preference-based reinforcement learning via preferred trajectory generation.
\newblock In \emph{The Twelfth International Conference on Learning Representations}, 2024.
\newblock URL \url{https://openreview.net/forum?id=EG68RSznLT}.

\bibitem[Zhu et~al.(2024)Zhu, Zhao, He, Zhong, Zhang, Guo, Chen, and Zhang]{zhu2024diffusionmodelsreinforcementlearning}
Zhu, Z., Zhao, H., He, H., Zhong, Y., Zhang, S., Guo, H., Chen, T., and Zhang, W.
\newblock Diffusion models for reinforcement learning: A survey, 2024.
\newblock URL \url{https://arxiv.org/abs/2311.01223}.

\bibitem[Ziegler et~al.(2019)Ziegler, Stiennon, Wu, Brown, Radford, Amodei, Christiano, and Irving]{ziegler2019fine}
Ziegler, D.~M., Stiennon, N., Wu, J., Brown, T.~B., Radford, A., Amodei, D., Christiano, P., and Irving, G.
\newblock Fine-tuning language models from human preferences.
\newblock \emph{arXiv preprint arXiv:1909.08593}, 2019.

\end{thebibliography}
\bibliographystyle{icml2025}

%%%%%%%%%%%%%%%%%%%%%%%%%%%%%%%%%%%%%%%%%%%%%%%%%%%%%%%%%%%%%%%%%%%%%%%%%%%%%%%
%%%%%%%%%%%%%%%%%%%%%%%%%%%%%%%%%%%%%%%%%%%%%%%%%%%%%%%%%%%%%%%%%%%%%%%%%%%%%%%
% APPENDIX
%%%%%%%%%%%%%%%%%%%%%%%%%%%%%%%%%%%%%%%%%%%%%%%%%%%%%%%%%%%%%%%%%%%%%%%%%%%%%%%
%%%%%%%%%%%%%%%%%%%%%%%%%%%%%%%%%%%%%%%%%%%%%%%%%%%%%%%%%%%%%%%%%%%%%%%%%%%%%%%
% \newpage
% \appendix
% \onecolumn
% \section{You \emph{can} have an appendix here.}

% You can have as much text here as you want. The main body must be at most $8$ pages long.
% For the final version, one more page can be added.
% If you want, you can use an appendix like this one.  

% The $\mathtt{\backslash onecolumn}$ command above can be kept in place if you prefer a one-column appendix, or can be removed if you prefer a two-column appendix.  Apart from this possible change, the style (font size, spacing, margins, page numbering, etc.) should be kept the same as the main body.
%%%%%%%%%%%%%%%%%%%%%%%%%%%%%%%%%%%%%%%%%%%%%%%%%%%%%%%%%%%%%%%%%%%%%%%%%%%%%%%
%%%%%%%%%%%%%%%%%%%%%%%%%%%%%%%%%%%%%%%%%%%%%%%%%%%%%%%%%%%%%%%%%%%%%%%%%%%%%%%

\end{document}